\documentclass{article}

\usepackage[a4paper, margin=1in]{geometry}

\usepackage[utf8]{inputenc} 
\usepackage[T1]{fontenc}    
\usepackage{hyperref}       
\usepackage{url}            
\usepackage{booktabs}       
\usepackage{multirow}
\usepackage{listings}
\usepackage{rotating} 
\usepackage{setspace}
\usepackage{wrapfig} 
\usepackage{amsfonts} 
\usepackage{amsmath}
\usepackage{graphicx} 
\DeclareGraphicsExtensions{.png,.pdf,.jpg,.jpeg}
\usepackage{nicefrac} 
\usepackage{microtype} 
\usepackage{xcolor} 
\usepackage{authblk}
\usepackage{caption}
\usepackage{subcaption}
\usepackage{natbib}
\usepackage{lineno}

\title{How well can off-the-shelf LLMs elucidate molecular structures from mass spectra using chain-of-thought reasoning?}

\author{
  Yufeng Wang$^{1,\dag}$,
  Wei Lu$^{1,\dag}$,
  Lin Liu$^{2}$,
  Hao Xu$^{3,*}$,
  Haibin Ling$^{1,*}$\\
  $^1$Department of Computer Science, Stony Brook University\\
  $^2$Department of Chemistry, Stanford University\\
  $^3$Department of Medicine, Brigham and Women's Hospital, Harvard Medical School\\
}

\begin{document}
\date{}
\maketitle

\renewcommand{\thefootnote}{\fnsymbol{footnote}}

\footnotetext[1]{Corresponding authors: \texttt{haibin.ling@gmail.com}, \texttt{haxu@bwh.harvard.edu}}
\footnotetext[2]{These authors contributed equally to this work}

\renewcommand{\thefootnote}{\arabic{footnote}}

\begin{abstract}
Mass spectrometry (MS) is a powerful analytical technique for identifying small molecules, yet determining complete molecular structures directly from tandem mass spectra (MS/MS) remains a long-standing challenge due to complex fragmentation patterns and the vast diversity of chemical space. 
Recent progress in large language models (LLMs) has shown promise for reasoning-intensive scientific tasks, but their capability for chemical interpretation is still unclear. 
In this work, we introduce a \textbf{Chain-of-Thought (CoT) prompting framework} and benchmark that evaluate how LLMs reason about mass spectral data to predict molecular structures. 
We formalize expert chemists' reasoning steps—such as double bond equivalent (DBE) analysis, neutral loss identification, and fragment assembly—into structured prompts and assess multiple state-of-the-art LLMs (Claude-3.5-Sonnet, GPT-4o-mini, and Llama-3 series) in a zero-shot setting using the MassSpecGym dataset. 
Our evaluation across metrics of SMILES validity, formula consistency, and structural similarity reveals that while LLMs can produce syntactically valid and partially plausible structures, they fail to achieve chemical accuracy or link reasoning to correct molecular predictions. 
These findings highlight both the interpretive potential and the current limitations of LLM-based reasoning for molecular elucidation, providing a foundation for future work that combines domain knowledge and reinforcement learning to achieve chemically grounded AI reasoning.


\end{abstract}

\section{Introduction}

Mass spectrometry (MS) is a fundamental analytical technique for small-molecule identification, widely used in natural product discovery, metabolomics, and drug development \cite{wolfender2018accelerating}. Its high sensitivity, fast throughput, and ability to resolve complex mixtures make it suitable for analyzing the chemical composition of biological and environmental samples \cite{patti2012metabolomics}. By providing accurate mass measurements and fragmentation patterns through tandem MS (MS/MS), this technique enables determination of molecular formulas and substructures, supporting both the identification of known compounds and the partial characterization of unknowns \cite{theodoridis2011mass}. Despite these advantages, determining complete molecular structures from MS data alone remains a long-standing challenge due to the ambiguity of fragmentation spectra and the vast diversity of chemical space \cite{stein2012mass}. Consequently, expert chemists rely on domain knowledge, careful interpretation of fragment ions, and iterative hypothesis refinement to infer plausible molecular structures.

Recently, artificial intelligence (AI) methods, especially deep generative and discriminative models, have been increasingly explored for molecular structure elucidation from mass spectrometry data \cite{duhrkop2019sirius, stravs2022msnovelist, shrivastava2021massgenie, litsa2023end, goldman2023annotating, bohde2025diffms, wang2025madgen, yang2024structural}. Early systems such as SIRIUS formulate identification as a fragmentation-tree optimization problem, using Bayesian scoring over putative fragment trees to infer molecular formulas and rank candidate structures from spectral libraries \cite{duhrkop2019sirius}. MSNovelist combines spectrum-derived molecular fingerprints with an encoder–decoder network that generates SMILES de novo from MS/MS data, enabling open-set structure prediction beyond database lookup \cite{stravs2022msnovelist, honda2019smiles, priyadarsini2024self}. MassGenie trains a large Transformer (around 400M parameters) directly on collections of simulated spectra and SMILES strings to learn mappings from fragmentation patterns to structures \cite{shrivastava2021massgenie}. End-to-end encoder–decoder models such as Spec2Mol \cite{litsa2023end, litsa2021spec2mol} similarly process spectra with a neural encoder and decode candidate SMILES sequences. These approaches perform well when spectra contain rich fragmentation information, but their performance often degrades on experimental MS/MS data due to weak generalization to unseen chemotypes and the lack of explicit domain constraints (e.g., atom valency or molecular formulas). In addition, sequence-based models are sensitive to peak order, can produce inconsistent results under minor spectral perturbations, and are generally difficult to interpret because they output structures without exposing intermediate chemical reasoning or substructure relationships.


\begin{figure}[t]
    \centering
    \includegraphics[width=\linewidth,keepaspectratio]{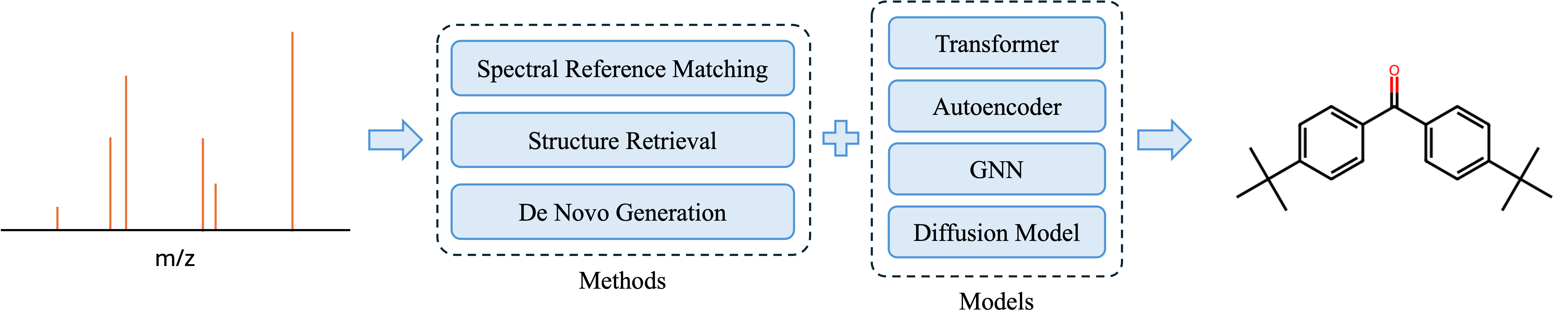}
    \caption{\textbf{Standard machine-learning pipeline for MS/MS-to-molecule prediction.} 
    Given an input MS/MS spectrum (left), existing methods typically combine
    (i) spectral reference matching (library search),
    (ii) structure retrieval from candidate databases, and
    (iii) de novo generation models using architectures such as Transformers,
    autoencoders, GNNs, or diffusion models,
    to propose one or more molecular structures (right).}
    \label{fig:ms2mol-workflow}
\end{figure}

\begin{figure}[h]
    \centering
    \captionsetup{justification=raggedright,singlelinecheck=false}

    \begin{subfigure}[b]{\linewidth}
        \centering
        \caption{The workflow of CoT-guided MS-to-SMILES benchmarking framework.}
        \vspace{2pt}
         \includegraphics[width=\linewidth]{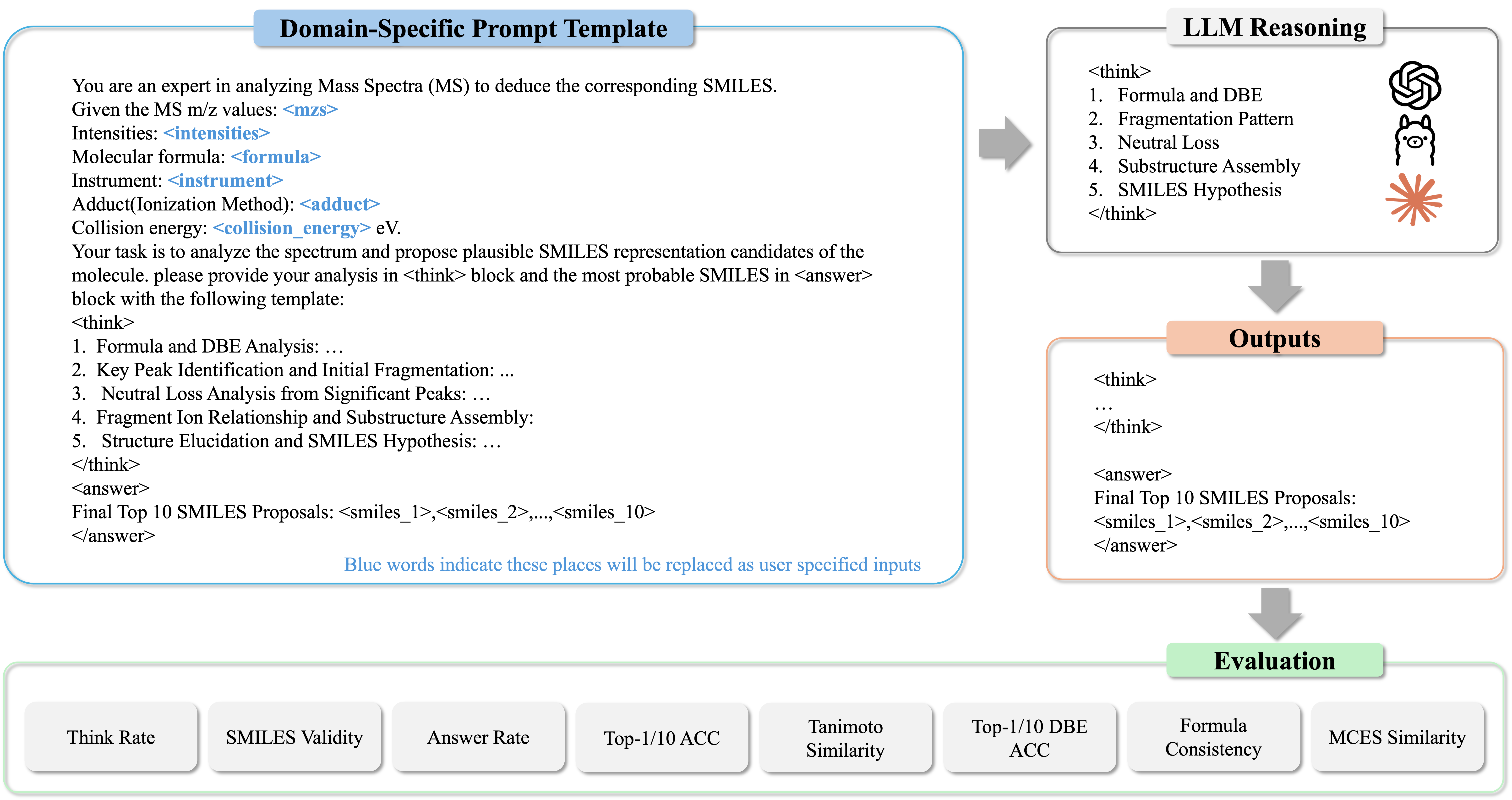}
    \end{subfigure}
    \vspace{3pt}
    
    \begin{subfigure}[b]{\linewidth}
        \centering
        \caption{Example prediction: ground-truth structure and LLM predicted 10 SMILES candidates.}
        \vspace{4pt}
        \includegraphics[width=\linewidth]{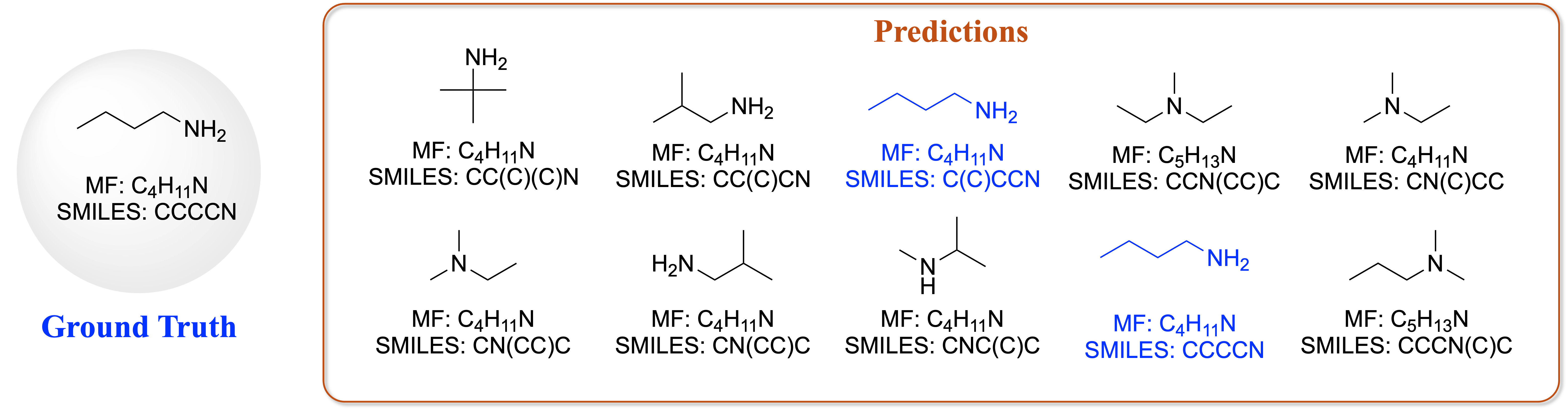}
    \end{subfigure}
    \caption{\textbf{Overview of the CoT-guided MS-to-SMILES benchmarking framework.} A domain-specific prompt elicits structured CoT reasoning from an LLM, which generates ranked SMILES candidates that are evaluated with format, validity, and structural/chemical metrics.}
    \label{fig:workflow}
\end{figure}


To overcome some of these limitations, more recent work has integrated domain-specific priors and structured chemical representations into spectrum-to-structure models \cite{chen2021molecular,chen2024molecular,goldman2023annotating,bohde2025diffms,wang2025madgen, le2020neuraldecipher}. MIST \cite{goldman2023annotating} proposes a spectrum-to-fingerprint Transformer that leverages peak annotations, neutral losses, and a progressive fingerprint decoding process, improving interpretability and downstream reconstruction. DiffMS \cite{bohde2025diffms} extends this idea with a formula-constrained graph diffusion model that generates molecular graphs directly from spectra while respecting atom-composition constraints; the diffusion mechanism also supports permutation invariance in both molecular and spectral representations. MADGEN \cite{wang2025madgen} introduces a two-stage approach that first retrieves a scaffold via contrastive learning and then generates full structures around it, narrowing the search space and improving robustness for sparse spectra. While these methods emphasize structural constraints and inductive biases, they remain primarily model-centric: they focus on learning accurate mappings from spectra to structures but still provide limited visibility into the reasoning process by which spectral evidence is translated into molecular hypotheses.

In parallel, LLMs have demonstrated strong general-purpose reasoning capabilities. CoT prompting encourages models to generate intermediate reasoning steps before producing an answer, and has been shown to substantially improve performance on tasks that require multi-step logic, arithmetic, and symbolic manipulation \cite{wei2022chain}. Beyond pure prompting, reinforcement learning has emerged as a powerful tool for sharpening these reasoning abilities. InstructGPT and related RLHF systems fine-tune LLMs with human preference feedback, yielding models that follow instructions more faithfully, provide more truthful explanations, and better align their step-by-step reasoning with user intent \cite{ouyang2022training, qiuphenomenal, lewkowycz2022solving, shao2024deepseekmath, li2023chain}. ReAct-style agents interleave natural-language reasoning with external actions, generating thought–action trajectories that query tools or environments and update plans based on observations \cite{yao2023react}, and more recent work explores RL-based curricula that construct harder reasoning traces for complex decision-making problems \cite{xu2025towards,yao2023tree}. Together, these CoT and RL-based advances move beyond static pattern matching toward models that can iteratively plan, reflect, and adapt, suggesting a promising, yet largely unexplored direction for applying LLM-style reasoning to physically grounded scientific domains such as mass spectral structure elucidation.

Interpreting MS/MS spectra to determine molecular structures is inherently a multi-step analytical process: chemists start from the precursor formula and double bond equivalent (DBE), identify key peaks and neutral losses, group fragments into substructures, and finally assemble a globally consistent molecular graph \cite{pei2023biot5, edwards2021text2mol}. These characteristics make the MS-to-molecule problem naturally suitable for CoT-style reasoning. However, it is not yet clear to what extent general-purpose LLMs, which are trained primarily on textual data and have never seen raw spectra, can perform such structured analysis or whether their explanations meaningfully reflect underlying chemical constraints \cite{ghugare2023searching}.

In this work, we explore whether LLMs can apply structured reasoning to the task of molecular structure prediction from mass spectra. We design a \textbf{CoT framework} that formalizes the logical process used by chemists, starting from formula and DBE analysis, identifying key peaks, examining neutral losses, connecting fragment ions, and finally proposing a consistent SMILES structure, and encode this process into a domain-specific prompt template for MS-to-SMILES prediction. Based on this framework, we build a benchmarking protocol that evaluates multiple LLMs on both their structural predictions and their reasoning behavior using the MassSpecGym dataset. Rather than proposing a new spectrum-trained generative model, our goal is to establish a reasoning-centered benchmark: a systematic CoT-based protocol for assessing how general-purpose LLMs interpret spectral information, articulate intermediate chemical reasoning, and where their behavior diverges from mechanistic expectations. This benchmark serves as a diagnostic tool that highlights the reasoning patterns and limitations of LLMs when applied to mass spectral structure elucidation, complementing prior model-centric work on spectrum-to-structure prediction.

\section{Methods}
\subsection{Problem Formulation as a Reasoning and Generation Task}
\label{sec:problem_formulation}

The goal of this work is to predict a molecular structure, represented as a SMILES string, from a given mass spectrum and the molecular formula. We cast this as a \textit{conditional generation} problem evaluated under standardized prompting (no model training).

\paragraph{Input Space}
An input instance $x$ consists of two primary components:
\begin{enumerate}
    \item The mass spectrum $S$: a set of $N$ peak pairs, $S=\{(\mathrm{m/z}_k,\, i_k)\}_{k=1}^{N}$, where $\mathrm{m/z}_k \in \mathbb{R}^+$ is the mass-to-charge ratio of the $k$-th peak and $i_k \in [0,1]$ is its normalized relative intensity.
    \item The molecular formula $F$: a string for the elemental composition of the precursor ion (e.g., ``C$_6$H$_{12}$O$_6$'').
\end{enumerate}
Thus, the input space is $\mathcal{X}=\{(S,F)\}$, optionally augmented in the prompt with instrument, adduct, and collision energy metadata.

\paragraph{Output Space}
For SMILES prediction, the output $y_{\text{SMILES}}$ is a string in the set $\mathcal{Y}_{\text{SMILES}}$ of valid canonical SMILES. In our benchmark, models \textbf{only} generate SMILES strings; we do not predict molecular fingerprints as outputs.

\paragraph{Task Definition}
Given $x=(S,F)$, a pre-trained LLM is prompted to produce $(C,\, \hat{y}_{\text{SMILES}})$, where $C\in\mathcal{C}$ is an optional CoT trace describing intermediate reasoning. Formally, we evaluate a prompted mapping
\begin{equation}
f': \mathcal{X} \longrightarrow \mathcal{C} \times \mathcal{Y}_{\text{SMILES}},
\end{equation}
with no parameter updates. The primary objective is the accuracy and plausibility of $\hat{y}_{\text{SMILES}}$; $C$ is analyzed qualitatively for reasoning fidelity.

\paragraph{Evaluation Protocol}
Models are \textit{prompted (zero-shot)} and evaluated on structure-centric metrics derived from the predicted SMILES:
(i) SMILES validity (parsability by RDKit),
(ii) formula consistency (atom counts match $F$),
(iii) DBE accuracy (computed from the predicted SMILES agrees with the DBE implied by $F$),
(iv) Top-$k$ exact match accuracy, and
(v) Top-$k$ structural similarity using Tanimoto on fingerprints \textit{computed from SMILES} and graph overlap via minimum graph dissimilarity (MCES).
We do not evaluate fingerprint prediction as a standalone task. CoT quality is assessed through format adherence and step-wise chemical plausibility checks described in the Experiments.

This formulation uses the generative abilities of LLMs to propose structures while exposing the intermediate reasoning via CoT.

\subsection{CoT Prompting}

To guide structure prediction from spectra, we use a CoT prompting strategy embedded in the LLM input. The reasoning chain follows five steps: 
\begin{itemize}
    \item \textbf{Formula and DBE analysis} to constrain feasible scaffolds;
    \item \textbf{Key peak identification} (molecular ion and base peak) to anchor molecular weight and stable fragments;
    \item \textbf{Neutral loss analysis} from prominent ions to infer functional groups;
    \item \textbf{Fragment-ion relationships and substructure assembly} to connect candidate substructures consistently with $F$ and DBE; and
    \item \textbf{Structure elucidation and SMILES hypothesis} that synthesizes the evidence into candidate structures.
\end{itemize}

The template provides explicit headings and brief instructions for each stage (e.g., DBE: C $-$ H/2 $+$ N/2 $+ 1$”), promoting a consistent, chemically meaningful reasoning order from global constraints to specific fragments. This design mirrors expert practice, improves transparency of the model’s analysis, and supports qualitative error analysis.


Here is an example LLM input for given specifications:

\begin{center}
  \begin{minipage}{0.9\linewidth}
      \begin{singlespace}   

    \begin{lstlisting}
MS m/z values: [53.0024, 53.9977, 55.0544, 56.0496, 57.07, 74.0965]
Intensities: [0.014, 0.003, 0.002, 0.002, 1.0, 0.434]
Molecular formula: C4H11N
Instrument: Orbitrap
Adduct: [M+H]+
Collision energy: 35.0 eV
    \end{lstlisting}
    \end{singlespace}
  \end{minipage}
\end{center}
And the example LLM output is:

\begin{center}
  \begin{minipage}{0.9\linewidth}
      \begin{singlespace}   

    \begin{lstlisting}
<think>
1. Formula and DBE Analysis:
    * Formula: C4H11N
    * Double Bond Equivalents (DBE): DBE = C - H/2 + N/2 + 1 = 4 - (11/2) + (1/2) + 1 = 4 - 5.5 + 0.5 + 1 = 0
    * Initial structural implications from DBE: DBE = 0 indicates a fully saturated compound, so no rings or double bonds. The formula is compatible with primary/secondary/tertiary alkylamines; specifically, straight-chain or branched but saturated.

2. Key Peak Identification and Initial Fragmentation:
    * Base Peak: m/z 57.07 with intensity 1.0. [M+H]+ calculated:  
      Monoisotopic mass C4H11N = (4*12.0000) + (11*1.0078) + (14.0031) = 48.0000 + 11.0858 + 14.0031 = 73.0889  
      [M+H]+ = 74.0967, matches m/z 74.0965 (intensity 0.434).  
      Thus, m/z 74.0965 is the protonated molecular ion.  
      m/z 57.07 (base peak) thus represents [M+H - 17]+ (= -17.03 Da).
      17 Da neutral loss corresponds to NH3 (ammonia), indicating facile elimination of ammonia typical for primary or secondary amines.

3. Neutral Loss Analysis from Significant Peaks:
    * From [M+H]+ (m/z 74.0965):  
      Loss of 17 Da (m/z 57.07): loss of NH3, leaving a C4 fragment, likely a butyl carbocation (C4H9+).
    * Lower m/z peaks:
        - m/z 56.0496 (relative intensity 0.002): difference from 57.07 is ~1 Da, consistent with loss of H (radical loss).
        - m/z 55.0544 (intensity 0.002): further 1 Da loss (H).
        - m/z 53.0024 (intensity 0.014): loss of 4 Da from m/z 57.07, possibly loss of 2x H2 or C2H4 (Ethene, 28 Da).
        - m/z 53.9977 (intensity 0.003): similar logic, but exact fragments less direct without more context.
    * Common neutral losses: NH3 (17 Da, base peak), H (1 Da), ethyl/alkyl fragments. All consistent with typical alkylamine fragmentation.

4. Fragment Ion Relationship and Substructure Assembly:
    * Peak at m/z 57.07: [M+H-NH3]+; the structure after simple ammonia loss from C4H11N is C4H9+ (likely rearranged, possibly t-butyl cation for tert-butylamine, or sec-butyl). 
    * Peaks at m/z 56.0496/55.0544: possible sequential loss of H, common in carbocation rearrangements or dehydration.
    * Major fragmentation via ammonia loss strongly suggests primary (RNH2) or secondary amine (R2NH).
    * All possible C4H11N isomers:  
       - n-Butylamine (CH3CH2CH2CH2NH2)
       - sec-Butylamine ((CH3)2CHCH2NH2)
       - tert-Butylamine ((CH3)3CNH2)
       - isobutylamine ((CH3)2CHCH2NH2) [note: sec- and iso- have the same formula]
       - diethylamine (CH3CH2NHCH2CH3; secondary amine)

5. Structure Elucidation and SMILES Hypothesis:
    * The very prominent ammonia loss (NH3, 17 Da) and intense base peak at C4H9+ are both very typical for primary amines, especially simple n-butyl, iso-butyl, sec-butyl, tert-butyl, or secondary diethylamine.
    * The presence of a strong m/z 57.07 peak (C4H9+) is especially indicative of tert-butyl moiety (t-butylamine), as the t-butyl cation is highly stabilized and generally gives a dominant peak.
    \end{lstlisting}
    \end{singlespace}
  \end{minipage}
\end{center}

\begin{center}
  \begin{minipage}{0.9\linewidth}
      \begin{singlespace}   
    \begin{lstlisting}
    * The molecular ion is moderately abundant (0.43 rel. intensity), supporting a branched structure (t-butyl cation formation).
    * Less likely for straight chains (n-butylamine) where fragmentation would favor m/z 30/44, not 57.
    * Diethylamine can yield C4H9+ via cleavage, but tends to also give strong ethyl iminium peaks in lower m/z, absent here.
</think>
<answer>
Final 10 SMILES Proposals: CC(C)(C)N,CC(C)CN,C(C)CCN,CCN(CC)C,CN(C)CC,CN(CC)C,NCC(C)C,CNC(C)C,CCCCN,CCCN(C)C
</answer>

    \end{lstlisting}
    \end{singlespace}
  \end{minipage}
\end{center}

\section{Experiments and Results}
\label{sec:experiment}

In this section, we describe our evaluation setup and present a systematic analysis
of how CoT prompting affects LLM behavior on the MS-to-SMILES
task. We first outline the dataset, models, and metrics, then analyze global
performance, size-stratified behavior by molecular weight, and the consistency
of the generated CoT reasoning.

\subsection{Experimental Setup}
\label{subsec:exp-setup}

\paragraph{Dataset.}
We evaluate all models on the MassSpecGym benchmark \cite{bushuiev2024massspecgym}, which contains over 231{,}000 high-quality MS/MS spectra paired with approximately 29{,}000 unique molecular structures. Following DiffMS \cite{bohde2025diffms}, we adopt a split that enforces a MCES $\geq 10$ between training and test molecules. This constraint avoids trivial overlap in the underlying graphs and yields a challenging, yet realistic, test bed for assessing whether LLMs can generalize to novel chemotypes rather than memorizing common scaffolds.

\paragraph{Models and prompting.}
We benchmark four representative LLMs:
Claude-3.5-Sonnet, GPT-4o-mini, Llama-3-70B-Instruct, and Llama-3-8B-Instruct.
All models are used in a strict zero-shot setting and receive the same structured
CoT input template described in Section \ref{sec:problem_formulation}. The prompt
encodes the spectrum, precursor formula, and instrument metadata, and explicitly
asks the model to reason within a \texttt{<think>} block before listing the final
SMILES candidates in an \texttt{<answer>} block. No model training, fine-tuning, or
task-specific adaptation is performed; performance therefore reflects the native
reasoning and generation capabilities of current LLMs under a consistent,
domain-aware prompting scheme.

\paragraph{Evaluation metrics.}
Given a ground-truth SMILES $s_i$ and a set of $k$ predictions
$\{\hat{s}_i^{(1)}, \dots, \hat{s}_i^{(k)}\}$ for spectrum $i$, we quantify performance
along three axes: syntactic validity, chemical consistency, and structural similarity.

\emph{SMILES validity} measures whether the generated sequences can be parsed by
RDKit into valid molecular graphs. If $N$ is the number of spectra and
$N_\text{valid}$ is the number of valid predictions,
\begin{equation}
    \text{SMILES Validity} = \frac{N_\text{valid}}{N}.
\end{equation}

\emph{DBE accuracy} assesses whether the DBE computed from the predicted SMILES matches that of the ground truth:
\begin{equation}
    \text{DBE Accuracy} = \frac{1}{N} \sum_{i=1}^{N}
    \bigl[ \text{DBE}(\hat{s}_i) = \text{DBE}(s_i) \bigr].
\end{equation}
We also evaluate \emph{formula consistency} as the fraction of predictions whose
atom counts match the provided molecular formula, optionally combined with basic
valency checks (``self-consistency'').

\emph{Top-$k$ exact-match accuracy} is defined as
\begin{equation}
    \text{Top-}k\ \text{Accuracy}
    = \frac{1}{N} \sum_{i=1}^{N}
      \bigl[ s_i \in \{\hat{s}_i^{(1)}, \dots, \hat{s}_i^{(k)}\} \bigr],
\end{equation}
and requires that at least one of the predicted SMILES is an exact canonical match.

To soften this stringent criterion, we use two graded \emph{structural similarity}
metrics. The first is \emph{Top-$k$ maximum Tanimoto similarity} between Morgan
fingerprints of the ground truth and the predictions:
\begin{equation}
    \text{Top-}k\ \text{MTS}
    = \frac{1}{N} \sum_{i=1}^{N}
      \max_{j=1,\dots,k}
      \mathrm{Tanimoto}\bigl( \mathrm{FP}(s_i), \mathrm{FP}(\hat{s}_i^{(j)}) \bigr).
\end{equation}
The second is a graph-based measure derived from the MCES. For each prediction $\hat{s}_i^{(j)}$ we compute the normalized edge overlap,
and report the \emph{Top-$k$ MCES}:
\begin{equation}
    \text{Top-}k\ \text{Min MCES}
    = \frac{1}{N} \sum_{i=1}^{N}
      \min_{j=1,\dots,k}
      \left[
        1 - \frac{|\mathrm{MCES}(G(s_i), G(\hat{s}_i^{(j)}))|}
                 {\max\bigl(|E(G(s_i))|, |E(G(\hat{s}_i^{(j)}))|\bigr)}
      \right],
\end{equation}
where $G(\cdot)$ is the molecular graph and $E(\cdot)$ its edge set. A lower MCES
value indicates higher graph-level similarity. Together, these metrics let us
distinguish between models that merely produce syntactically valid SMILES and
those that capture deeper chemical and structural constraints.

\subsection{Global Quantitative Results}
\label{subsec:global-results}

\begin{table}[h]
\centering
\small
\begin{tabular}{@{}llcccc@{}}
\toprule
\textbf{Metric} & &
\textbf{Claude-3.5-Sonnet} &
\textbf{GPT-4o-mini} &
\textbf{Llama-3-8B} &
\textbf{Llama-3-70B} \\
\midrule
Think Rate (\%)  &        & 100.00 & 97.07 & 68.34 & 97.83 \\
Answer Rate (\%)              &        & 99.96  & 81.17 &  2.21 & 51.14 \\
SMILES Validity (\%)          &        & 93.70  & 75.41 &  1.44 & 38.12 \\
DBE Accuracy (\%)             &        &  3.94  &  3.91 &  4.22 &  4.10 \\
Formula Consistency (\%)      &        &  0.00  &  0.02 &  0.00 &  0.02 \\
\midrule
\multirow{2}{*}{Accuracy (\%)} 
                              & Top-1  & 0.00   & 0.00  & 0.00  & 0.00 \\
                              & Top-10 & 0.00   & 0.00  & 0.00  & 0.00 \\
\midrule
\multirow{2}{*}{\shortstack{Tanimoto\\Similarity}}
                              & Top-1  & 0.09   & 0.09  & 0.09  & 0.10 \\
                              & Top-10 & 0.10   & 0.12  & 0.10  & 0.12 \\
\midrule
\multirow{2}{*}{\shortstack{MCES}}
                              & Top-1  & 0.15   & 0.15  & 0.17  & 0.17 \\
                              & Top-10 & 0.14   & 0.07  & 0.16  & 0.14 \\
\bottomrule
\end{tabular}
\caption{\textbf{Performance of four LLMs on MS-to-SMILES prediction.}
Think/Answer rates measure CoT-format adherence; remaining metrics report syntactic validity,
chemical correctness, and structural similarity.}
\label{tab:results}
\end{table}

Table \ref{tab:results} summarizes global performance across the entire test
set. We highlight three main observations.

First, all models are capable of following the high-level CoT instructions, but
with varying reliability. Claude-3.5-Sonnet achieves nearly perfect adherence to
the prescribed format, with Think and Answer rates above 99.9\% and a SMILES
validity of 93.7\%. GPT-4o-mini also respects the \texttt{<think>} / \texttt{<answer>}
structure in most cases and yields valid SMILES in roughly three-quarters of
instances. In contrast, Llama-3-8B frequently truncates or misformats its output
(Answer Rate 2.21\%), which directly translates into a very low validity rate
(1.44\%). Scaling to Llama-3-70B substantially improves format adherence and
validity (51.14\% Answer Rate; 38.12\% Validity), indicating that larger models
are more robust under long, structured prompts.

Second, structural similarity remains modest even for the best-performing
models. Tanimoto scores hover around 0.09--0.12 and MCES values indicate only
partial overlap in graph connectivity. The open Llama models achieve slightly
higher MCES (up to 0.17 Top-1) than the proprietary models, despite
their lower validity; when Llama produces valid SMILES, those structures can
capture somewhat more of the target connectivity pattern. However, the
differences are small in absolute terms, and no model comes close to reproducing
the full scaffold or substitution pattern of the ground truth.

Third, chemical consistency metrics are uniformly poor. DBE accuracy remains in
the 3.9--4.2\% range, and formula consistency is effectively zero for all models
(except for a marginal 0.02\% for GPT-4o-mini and Llama-3-70B). Most strikingly,
Top-1 and Top-10 exact-match accuracy are identically zero across the board:
none of the models ever predicts the correct canonical SMILES, even when allowed
ten attempts. This shows that, although CoT prompting leads to outputs that
\emph{look} chemically sophisticated at the text level (e.g., DBE formulas,
named neutral losses), the final structures are neither formula-faithful
nor spectrally grounded.

Taken together, these results indicate that current LLMs can follow a structured,
chemically flavored reasoning template and generate syntactically valid SMILES
for some fraction of spectra, but they fail to translate that reasoning into
accurate or chemically self-consistent molecular structures. In the next
subsection, we ask whether this failure is uniform across the chemical space, or
whether certain molecule classes are more amenable to CoT-guided prediction.

\subsection{Size-Stratified Performance by Molecular Weight}
\label{subsec:binned-mw}

\begin{figure}[h]
    \centering
    \begin{subfigure}[b]{0.8\linewidth}
        \centering
        \caption{Top-1 Tanimoto similarity across molecular-weight bins for all four evaluated LLMs.}
        \includegraphics[width=\linewidth]{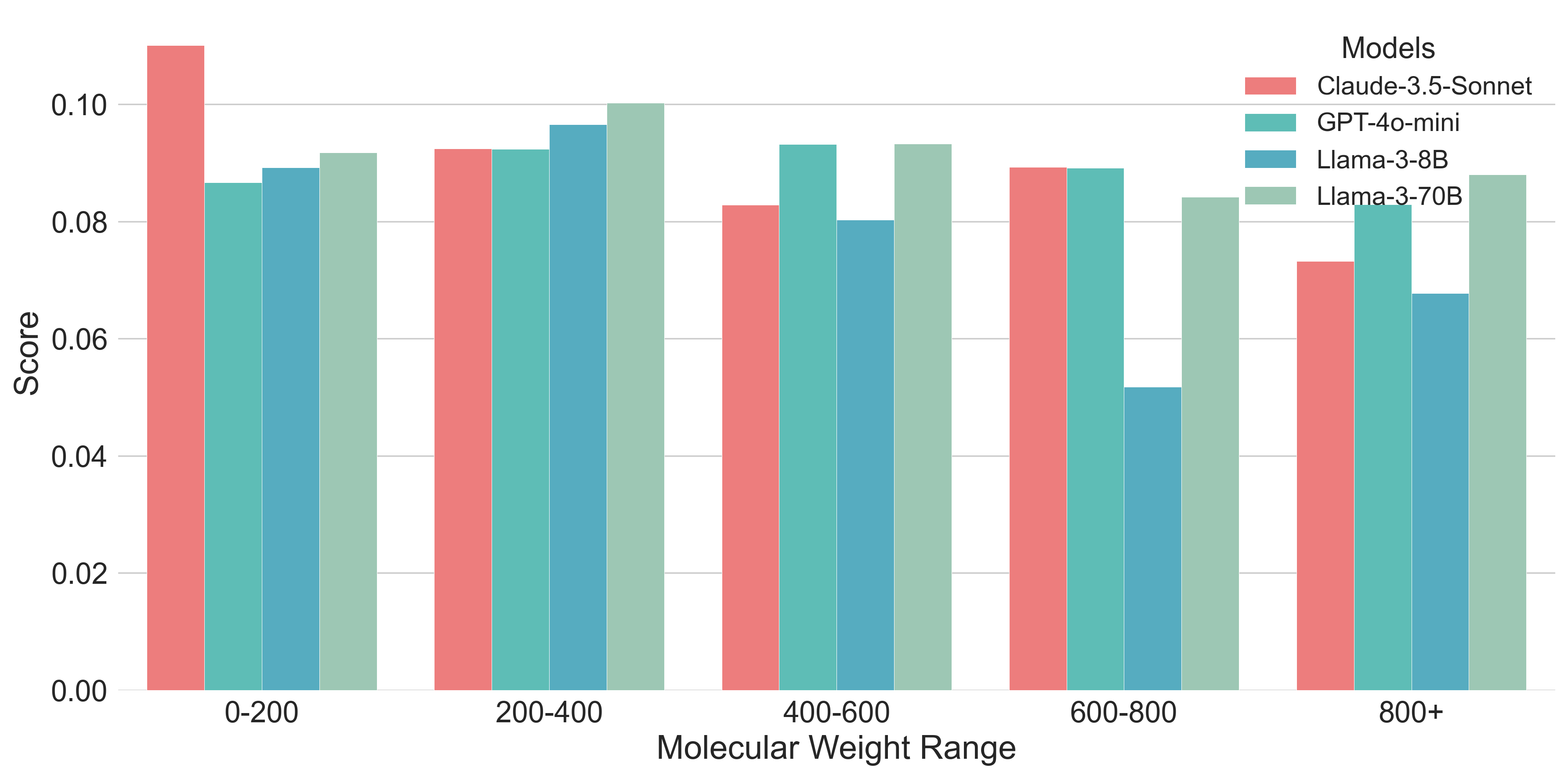}
        \label{fig:binned-tanimoto-top1}
    \end{subfigure}\vfill
    \begin{subfigure}[b]{0.8\linewidth}
        \centering
        \caption{Top-10 Tanimoto similarity across molecular-weight bins for all four evaluated LLMs.}
        \includegraphics[width=\linewidth]{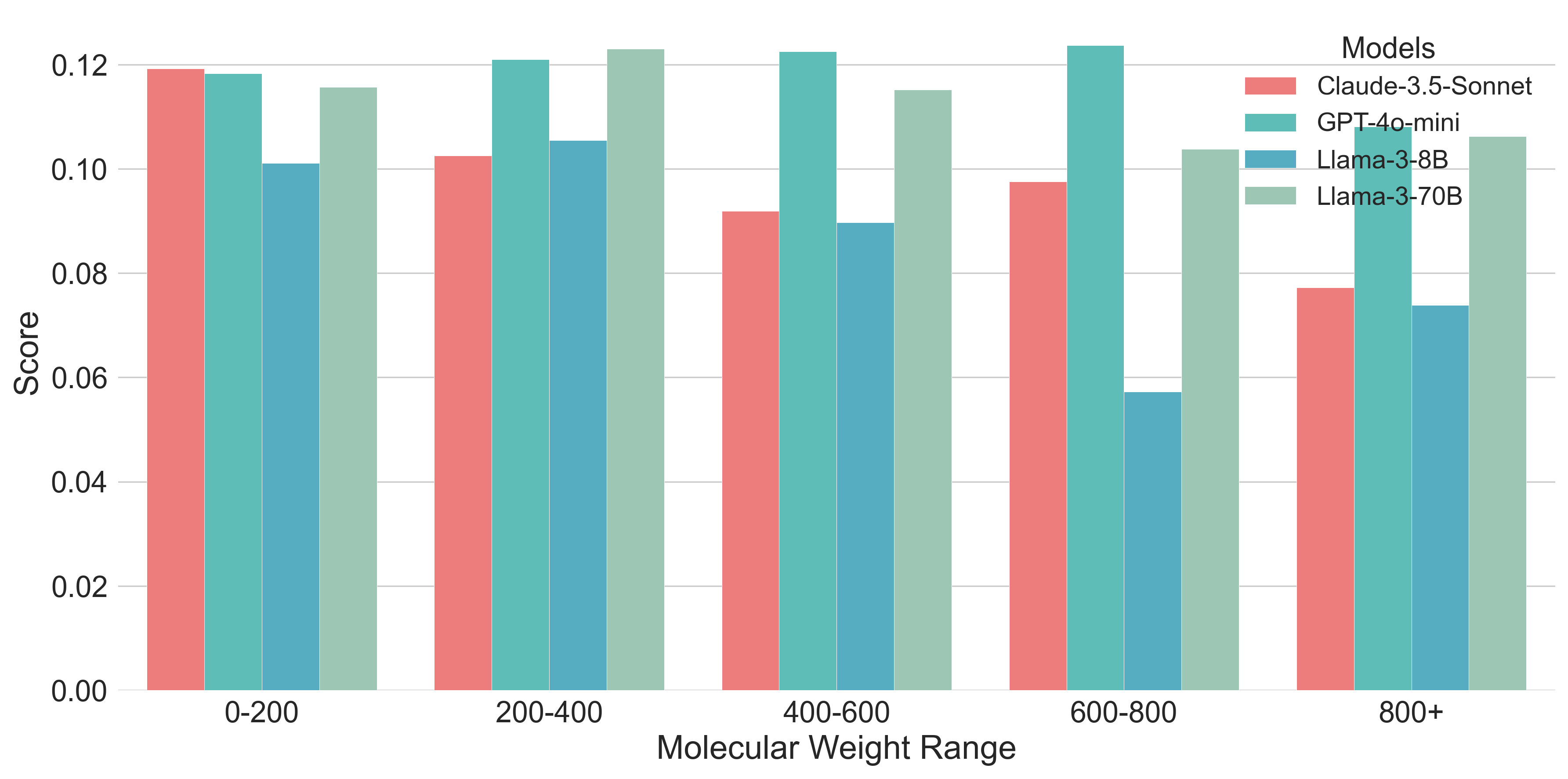}
        \label{fig:binned-tanimoto-top10}
    \end{subfigure}
    \caption{\textbf{Tanimoto similarity across molecular-weight bins for all four evaluated LLMs.} All models achieve their highest similarity in the 200-400 Da range, with performance degrading for very small ($<200$ Da) and larger ($>400$ Da) molecules.}
    \label{fig:binned-tanimoto}
\end{figure}

\begin{figure}[!h]
    \centering
    \begin{subfigure}[b]{0.8\linewidth}
        \centering
        \caption{Top-1 MCES across molecular-weight bins for all four evaluated LLMs.}
        \includegraphics[width=\linewidth]{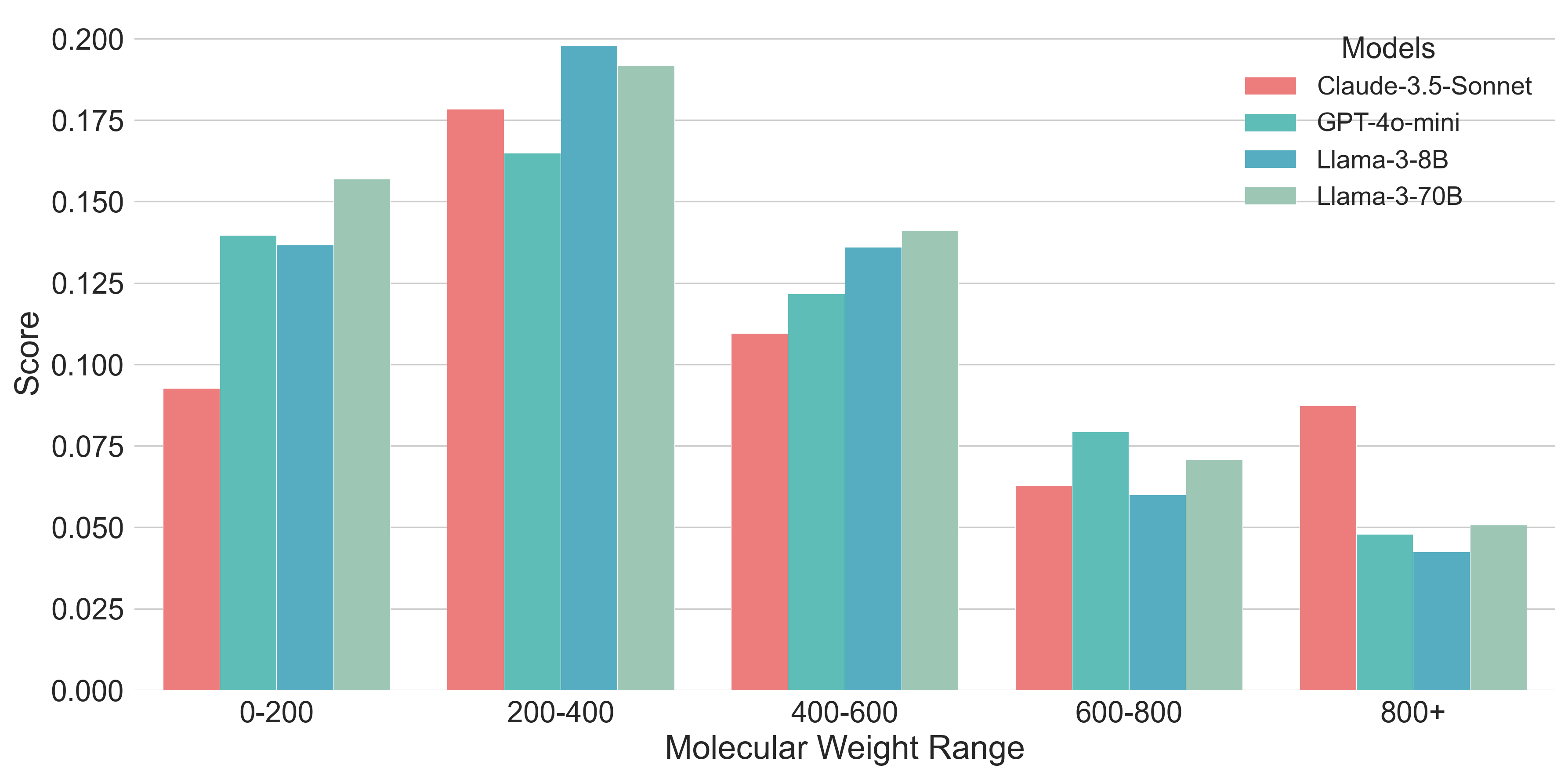}
        \label{fig:binned-mces-top1}
    \end{subfigure}\vfill
    \begin{subfigure}[b]{0.8\linewidth}
        \centering
        \caption{Top-10 MCES across molecular-weight bins for all four evaluated LLMs.}
        \includegraphics[width=\linewidth]{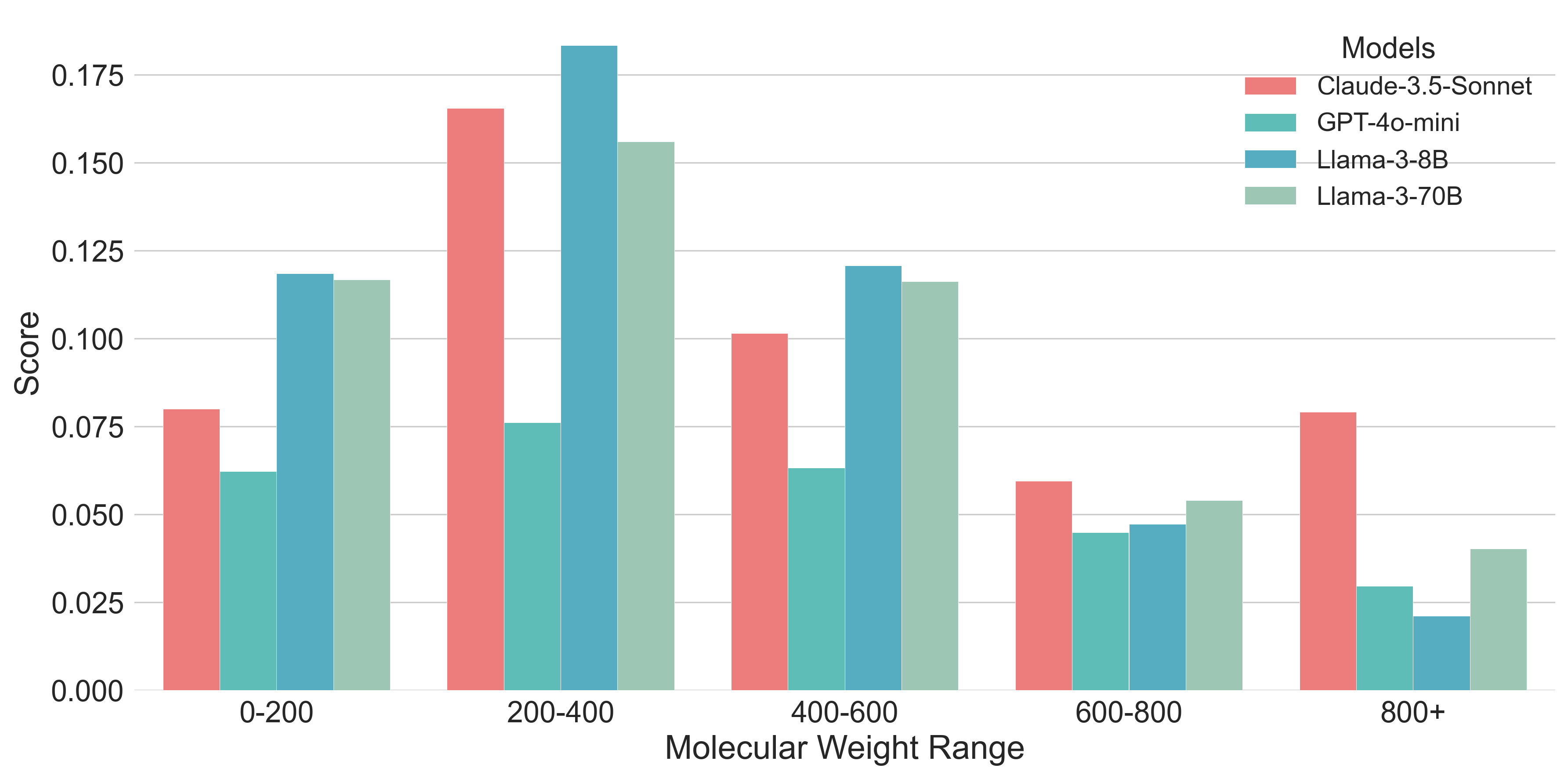}
        \label{fig:binned-mces-top10}
    \end{subfigure}
    \caption{\textbf{MCES across molecular-weight bins for all four evaluated LLMs.} MCES scores follow a similar trend as Tanimoto but exhibit an even sharper decline for large molecules ($>400$ Da), indicating severe mismatches in global graph connectivity.}
    \label{fig:binned-mces}
\end{figure}

To examine how molecular complexity influences LLM behavior, we stratify the MassSpecGym benchmark into five precursor-weight bins: $[0,200)$, $[200,400)$, $[400,600)$, $[600,800)$, and $[800,\infty)$ Da (Dalton, the standard molecular-mass unit \cite{newell2019international}). For each bin, we compute Top-1/Top-10 Tanimoto and MCES as well as Top-1/Top-10 exact-match accuracy across all four evaluated models (Claude-3.5-Sonnet, GPT-4o-mini, Llama-3-8B, and Llama-3-70B). Consistently, Top-1 and Top-10 exact-match accuracy remain zero in every bin and for every model, confirming that LLMs never reconstruct the correct SMILES in any molecular-weight regime; the absence of exact matches in global results is therefore not attributable to a few difficult molecules but reflects a pervasive inability to invert mass spectra into structures. More informative differences only appear when examining graded similarity metrics such as Tanimoto and MCES. Here, a consistent pattern emerges: structural similarity peaks in the 200-400 Da range. Claude-3.5-Sonnet achieves Top-1/10 Tanimoto scores of 0.092/0.103 and MCES of 0.178/0.166, GPT-4o-mini shows similar values (0.092/0.121 Tanimoto; 0.165/0.076 MCES), and the Llama models slightly exceed these numbers (e.g., Llama-3-8B at 0.097/0.105 Tanimoto and 0.198/0.183 MCES; Llama-3-70B at 0.100/0.123 Tanimoto and 0.192/0.156 MCES). These scores match or slightly surpass each model’s global averages, suggesting that mid-sized molecules represent a shared ``comfort zone'' where LLMs can assemble partially correct fragments—likely reflecting both their prevalence in the dataset and their overrepresentation in chemical corpora used during pretraining.

Beyond 400 Da, performance declines steadily for all models. For example, Llama-3-8B’s Top-1 Tanimoto drops from $\sim$0.10 in the 200-400 Da bin to $\sim$0.08 (400--600 Da) and $\sim$0.05 (600--800 Da), with MCES showing an even sharper decline; Llama-3-70B, Claude-3.5-Sonnet, and GPT-4o-mini exhibit comparable trajectories. This deterioration reflects the combinatorial explosion of chemically plausible structures at higher molecular weights: without any spectral encoder, chemical constraints, or modeling of fragmentation physics, LLMs effectively sample from a vast, weakly constrained scaffold space. Larger molecules also produce denser, more ambiguous fragment spectra, requiring mechanistic knowledge that current LLMs do not encode. Interestingly, very small molecules ($<200$ Da) also underperform compared to the 200-400 Da range. Although their structures are simpler, their spectra often contain fewer informative fragments, and many molecules share similar low-mass signatures; this intrinsic ambiguity is reflected in the modest Top-1 Tanimoto scores (0.087--0.092) and MCES values (0.14--0.16) across models in this bin. Sample distribution further amplifies these effects: the 200-400 Da bin dominates the dataset (16k--18k spectra for GPT-4o-mini and the Llama models), whereas the 800+ Da bin contains far fewer samples (hundreds per model; only 30 for Claude), so global averages are heavily influenced by the mid-mass region and partially obscure the severity of performance degradation at the extremes. Figures \ref{fig:binned-tanimoto} and \ref{fig:binned-mces} visualize these size-dependent trends across models and molecular-weight regimes.

Overall, the size-stratified analysis shows that CoT-guided LLMs do not simply ``fail more’’ on larger molecules. Instead, they exhibit a narrow region of partial plausibility for mid-sized molecules and degrade sharply on both sides: large molecules suffer from combinatorial structural complexity, while very small molecules lack sufficient spectral evidence. Yet even in the most favorable regime, similarity scores remain low and exact matches never occur, highlighting that CoT alone cannot bridge the gap between textual reasoning and the mechanistic interpretation required for MS-based structure elucidation.

\subsection{Qualitative Behavior of CoT Reasoning}
\label{subsec:cot-behavior}

\begin{table}[h]
\centering
\begin{tabular}{@{}lcccc@{}}
\toprule
\textbf{Model} &
\shortstack{\textbf{Avg CoT}\\\textbf{Length (words)}} &
\shortstack{\textbf{DBE}\\\textbf{Correctness (\%)}} &
\shortstack{\textbf{Formula}\\\textbf{Correctness (\%)}} &
\shortstack{\textbf{Contradiction}\\\textbf{Rate (\%)}} \\
\midrule
Claude-3.5-Sonnet & 214.1 & 3.9 & 0.1 & 72.8 \\
GPT-4o-mini       & 331.2 & 3.9 & 0.0 & 92.2 \\
Llama-3-8B        & 303.4 & 4.2 & 0.0 & 92.0 \\
Llama-3-70B       & 435.5 & 4.1 & 0.0 & 82.3 \\
\bottomrule
\end{tabular}
\caption{\textbf{CoT analysis across four LLMs.} DBE correctness measures whether the model
computes the correct double-bond equivalent in its reasoning; formula correctness
checks correctness of atom counts stated in the CoT; contradiction rate reflects
inconsistencies between CoT reasoning and the final generated SMILES. All values
are reported as percentages.}
\label{tab:cot-analysis}
\end{table}

While the previous sections focused on structural and chemical correctness of the final SMILES predictions, we also analyze the internal CoT reasoning to assess whether the step-by-step explanations contain chemically meaningful deductions. We quantify CoT behavior along four axes: DBE correctness, formula correctness, contradiction rate between the reasoning text and the final SMILES, and the average CoT length, which reflects verbosity rather than accuracy. Across all models, DBE correctness remains extremely low (3.9--4.2\%), showing that although CoT narratives often include DBE calculations, they almost never match the true DBE of the target molecule. Formula correctness is essentially zero (0.1\% for Claude-3.5-Sonnet and 0.0\% for the remaining models), indicating that models fail to maintain accurate atom counts during the reasoning process. These errors occur even when the reasoning text appears chemically styled, confirming that such explanations are largely ungrounded and do not reflect true chemical computation.

The contradiction rate further highlights this disconnect between linguistic reasoning and chemical validity. Conflicts between the CoT and the final SMILES occur in 72.8\% of cases for Claude-3.5-Sonnet, 92.2\% for GPT-4o-mini, 92.0\% for Llama-3-8B, and 82.3\% for Llama-3-70B. Manual inspection shows that these contradictions often take the form of CoT descriptions of rings, DBE values, or functional groups that never appear in the final structure, or internally inconsistent statements within the same reasoning trace. Notably, larger models such as Llama-3-70B produce substantially longer CoT explanations on average (435.5 words), yet this increased verbosity does not improve DBE or formula correctness; instead, it often amplifies the amount of erroneous or contradictory reasoning produced. Together, these trends demonstrate that CoT explanations lack mechanistic grounding and fail to guide the model toward chemically valid outcomes.

Overall, the CoT analysis mirrors the structural-similarity findings: LLMs can fluently emulate the \emph{language} of expert mass spectral interpretation without internalizing the mechanistic principles required for accurate structure elucidation. In this setting, CoT prompting acts more as a stylistic scaffold than as a functional reasoning mechanism. Although the explanations appear detailed and expert-like, they rarely contain logically coherent or chemically correct deductions, and they frequently contradict the generated structures. This gap highlights the central challenge of applying text-only LLMs to scientific reasoning tasks where correctness depends on underlying physical and chemical constraints rather than linguistic plausibility.

\section{Discussion}
\label{sec:discussion}

The results presented in Section \ref{sec:experiment} collectively reveal a consistent and robust picture of how contemporary LLMs behave when tasked with MS-to-SMILES structure elucidation under CoT prompting. Despite their strong fluency in natural language and apparent ability to mimic expert-style reasoning patterns, all four evaluated models: Claude-3.5-Sonnet, GPT-4o-mini, Llama-3-8B, and Llama-3-70B, exhibit fundamental limitations in chemical fidelity, spectral grounding, and structural correctness. CoT prompting successfully enforces a structured output format (with Think/Answer adherence exceeding 97\% for Claude-3.5-Sonnet, GPT-4o-mini, and Llama-3-70B), yet this surface-level compliance does not translate into chemically valid reasoning: DBE accuracy remains between 3.9\% and 4.2\%, formula correctness is effectively zero, and none of the models ever produce the correct SMILES under Top-1 or Top-10 evaluation. CoT therefore acts primarily as a stylistic scaffold that shapes how models present their reasoning, not as a mechanism for grounding their predictions in the underlying chemistry of mass spectrometry.

\begin{table}[h]
\centering

\begin{tabular}{@{}lcccc@{}}
\toprule
\multirow{2}{*}{\textbf{Model}} &
\multicolumn{2}{c}{\textbf{Top-1}} &
\multicolumn{2}{c}{\textbf{Top-10}} \\
\cmidrule(lr){2-5}
& \textbf{Acc (\%)} & \textbf{Tanimoto} &
  \textbf{Acc (\%)} & \textbf{Tanimoto} \\
\midrule
\multicolumn{5}{c}{\textit{MassSpecGym baselines}} \\
\midrule
SMILES Transformer         & 0.00 & 0.03 & 0.00 & 0.10 \\
MIST + MSNovelist          & 0.00 & 0.06 & 0.00 & 0.15 \\
SELFIES Transformer        & 0.00 & 0.08 & 0.00 & 0.13 \\
Spec2Mol                   & 0.00 & 0.12 & 0.00 & 0.16 \\
MIST + NeuralDecipher      & 0.00 & 0.14 & 0.00 & 0.16 \\
Random Generation          & 0.00 & 0.08 & 0.00 & 0.11 \\
MADGEN                     & 0.80 & ---  & 1.60 & ---  \\
\textbf{DiffMS (SOTA)}     & \textbf{2.30} & \textbf{0.28} &
                             \textbf{4.25} & \textbf{0.39} \\
\midrule
\multicolumn{5}{c}{\textit{Zero-shot CoT-guided LLMs (this work)}} \\
\midrule
Claude-3.5-Sonnet          & 0.00 & 0.09 & 0.00 & 0.10 \\
GPT-4o-mini                & 0.00 & 0.09 & 0.00 & 0.12 \\
Llama-3-8B                 & 0.00 & 0.09 & 0.00 & 0.10 \\
Llama-3-70B                & 0.00 & 0.10 & 0.00 & 0.12 \\
\bottomrule
\end{tabular}
\caption{\textbf{Comparison of MassSpecGym structural prediction performance for prior MS-based models and CoT-guided LLMs.} Baseline numbers are taken from Bohde et al.\cite{bohde2025diffms}, they are all small models rather than LLMs. Our LLMs are evaluated under the same MassSpecGym split but in a strictly zero-shot setting with no spectral training.}
\label{tab:sota-massspecgym-llm}
\end{table}

When we compare LLMs to specialized MS-based structure generation systems, the limitations become even clearer. As shown in Table \ref{tab:sota-massspecgym-llm}, spectrum-trained architectures, including SMILES/SELFIES Transformers, Spec2Mol, MIST+NeuralDecipher, and especially DiffMS outperform LLMs by a substantial margin. DiffMS achieves 2.30\% Top-1 and 4.25\% Top-10 accuracy, with Tanimoto similarity reaching 0.39, three to four times higher than the best LLM scores (0.09–0.12). Even older baselines without diffusion, such as Spec2Mol or MIST+NeuralDecipher, match or exceed LLM performance despite being far smaller and domain-specific. This gap highlights a key distinction: while DiffMS and related models ingest real spectral features, enforce chemical constraints, and incorporate fragmentation physics, the LLMs in our study treat mass spectra as formatted text and rely solely on broad chemical patterns learned from language pretraining. Their predictions therefore reflect linguistic familiarity rather than mechanistic spectral interpretation.

\begin{figure}[!h]
    \centering
    
    \begin{subfigure}[t]{\linewidth}
        \centering
        \begin{minipage}[c]{0.2\linewidth}
            \centering
            \includegraphics[width=0.6\linewidth]{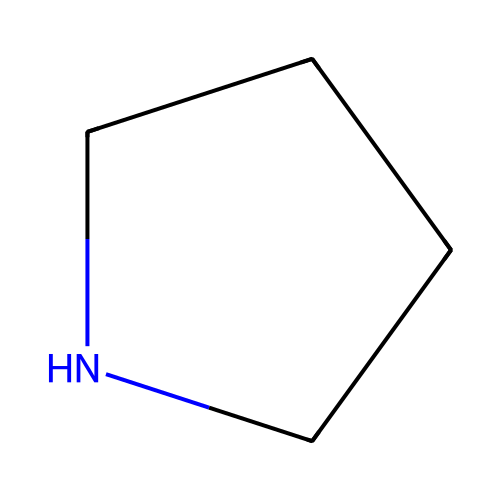}
        \end{minipage}\hfill
        \begin{minipage}[c]{0.8\linewidth}
            \centering
            \includegraphics[width=0.7\linewidth]{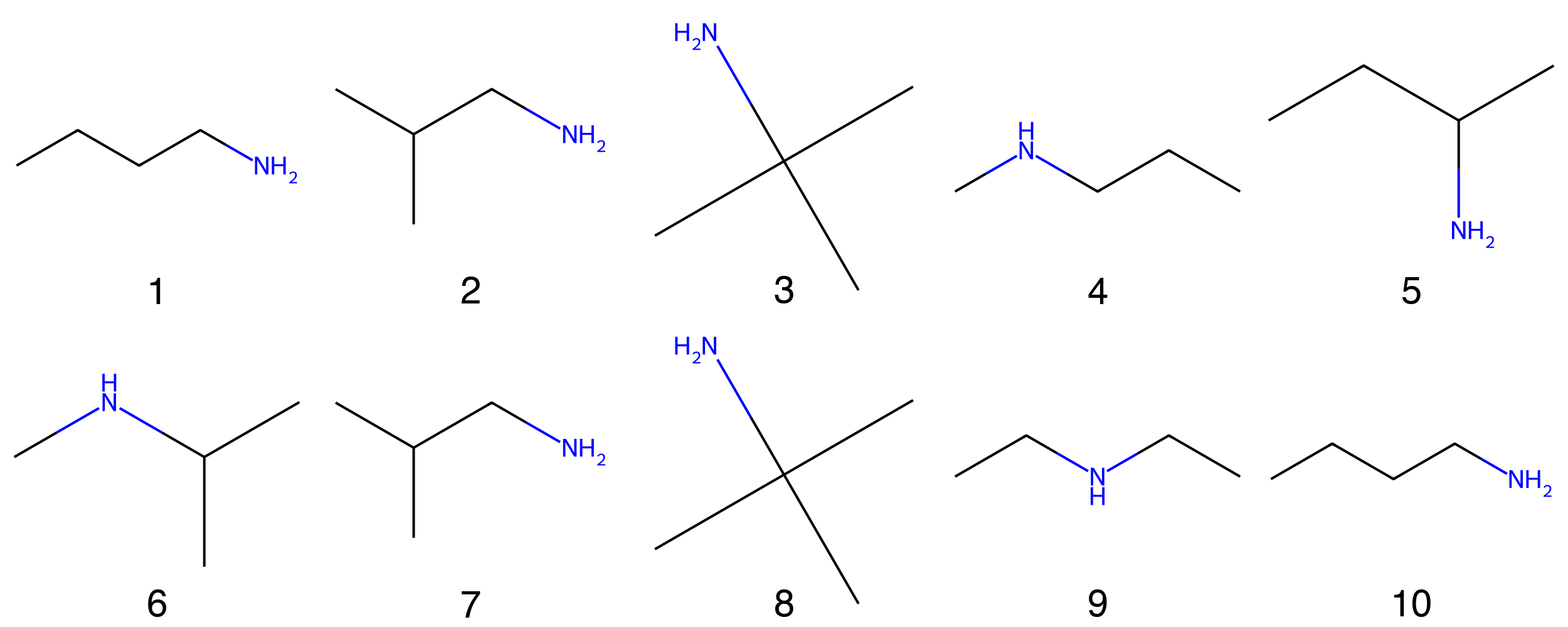}
        \end{minipage}
        \caption{Small molecule example ($<200$~Da).
        \textbf{Left:} ground-truth molecule.
        \textbf{Right:} 10 predicted structures. 
        All candidates have very low similarity to the reference
        (maximum Tanimoto $\approx 0.07$), illustrating near-random
        behavior in this regime.}
        \label{fig:case-small}
    \end{subfigure}
    
    \begin{subfigure}[t]{\linewidth}
        \centering
        \begin{minipage}[c]{0.2\linewidth}
            \centering
            \includegraphics[width=\linewidth]{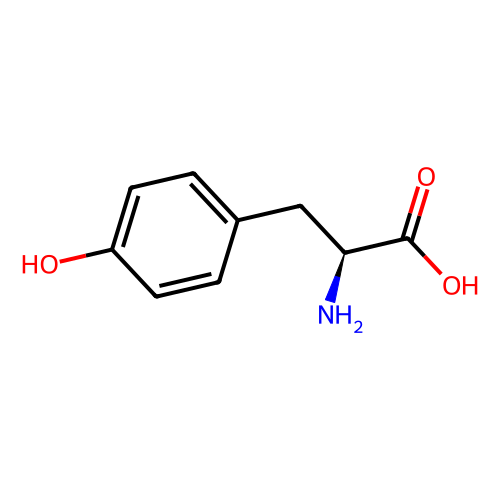}
        \end{minipage}\hfill
        \begin{minipage}[c]{0.79\linewidth}
            \centering
            \includegraphics[width=\linewidth]{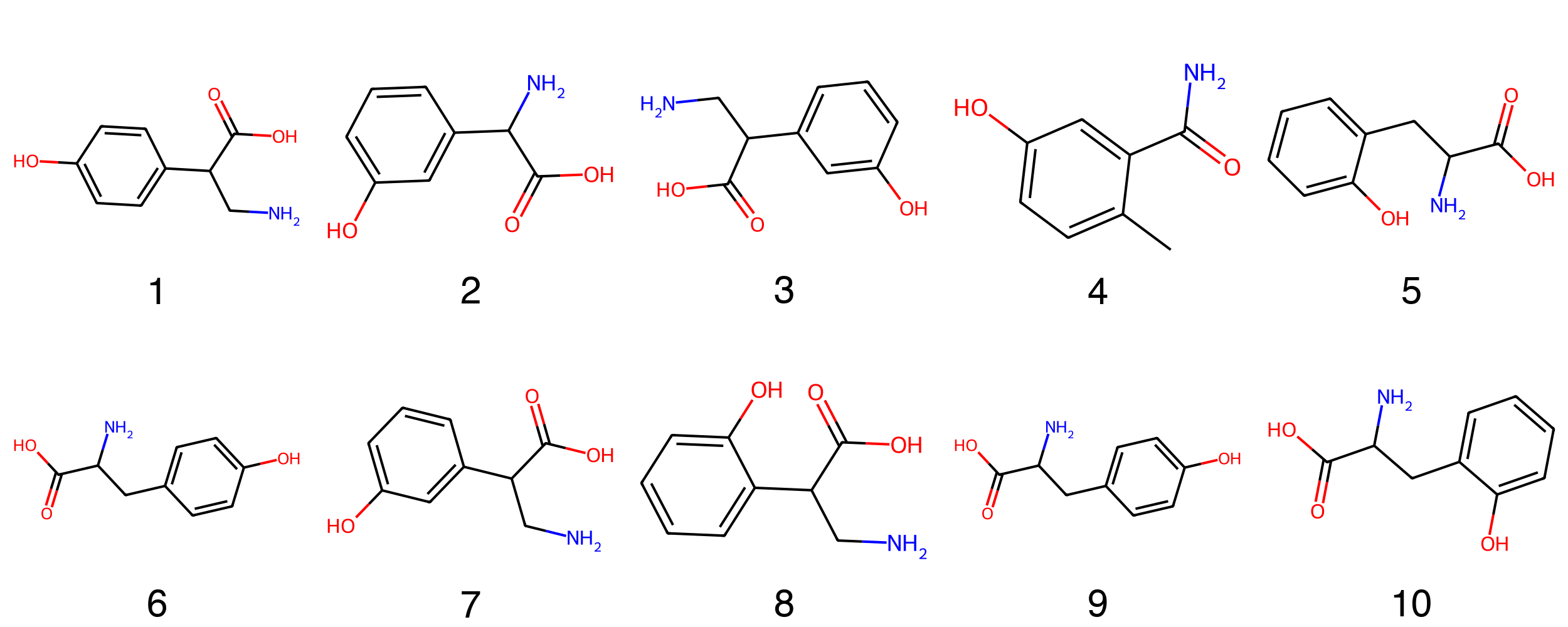}
        \end{minipage}
        \caption{Medium molecule example ($200-400$~Da). 
        \textbf{Left:} ground-truth molecule.
        \textbf{Right:} 10 predicted structures.
        Several candidates are close to the reference and exact matches
        (Tanimoto = 1.0) appear in the list, but not at rank~1, highlighting
        weak calibration of the model’s implicit scoring.}
        \label{fig:case-medium}
    \end{subfigure}

    \begin{subfigure}[t]{\linewidth}
        \centering
        \begin{minipage}[c]{0.2\linewidth}
            \centering
            \includegraphics[width=0.8\linewidth]{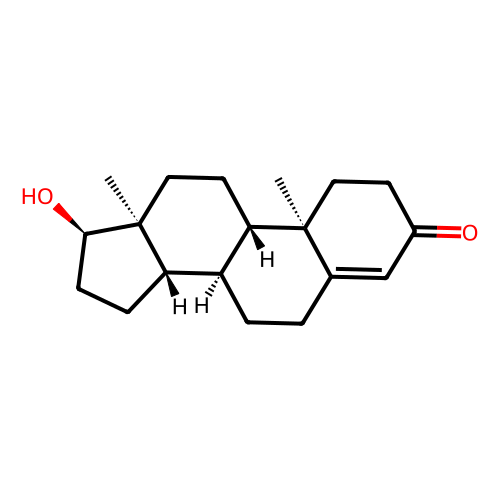}
        \end{minipage}\hfill
        \begin{minipage}[c]{0.8\linewidth}
            \centering
            \includegraphics[width=\linewidth]{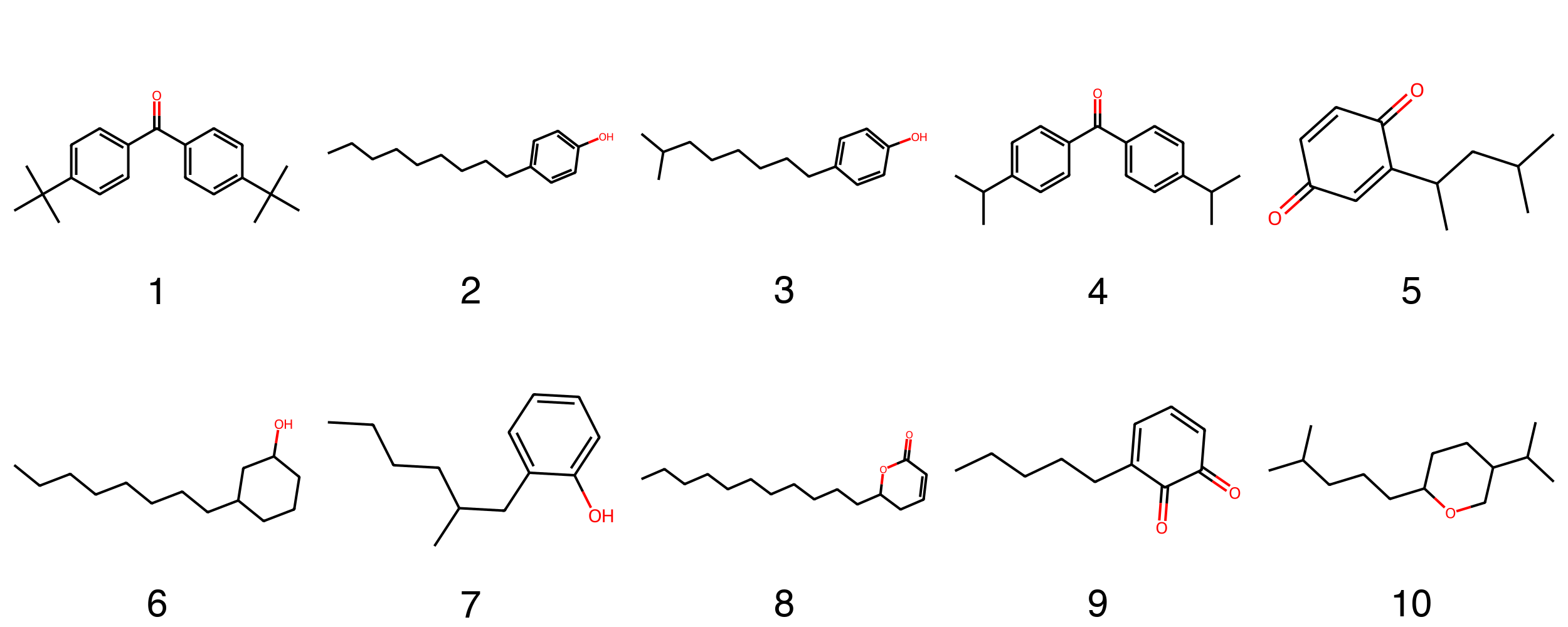}
        \end{minipage}
        \caption{Large molecule example ($>400$~Da).
        \textbf{Left:} ground-truth molecule.
        \textbf{Right:} 10 predicted structures. 
        All candidates remain far from the reference (maximum
        Tanimoto $\approx 0.13$), illustrating how the combinatorial
        explosion of high-mass structures leads to uniformly low
        similarity even among sampled predictions.}
        \label{fig:case-large}
    \end{subfigure}
    
    \caption{\textbf{Qualitative case studies across molecular-weight regimes.} Each row shows the model’s 10 candidate structures and the corresponding ground-truth molecule for a representative small, medium, and large precursor.}
    \label{fig:three-size-case-studies}
\end{figure}

The size-stratified analyses reinforce this interpretation. All four models perform best within the 200–400 Da range, likely due to a combination of dataset distribution (this bin is the most populated) and the prevalence of drug-like fragments in chemical text corpora. Yet even in this ``comfort zone,'' Top-1 Tanimoto remains roughly 0.10, far below the performance of spectrum-conditioned models. Beyond this range, performance decays sharply: 
small molecules ($<200$ Da) lack sufficient spectral fragment information to constrain the inverse problem (Fig. \ref{fig:three-size-case-studies}a), while
large molecules ($>400$ Da) become increasingly difficult due to the combinatorial explosion of plausible structural isomers (Fig. \ref{fig:three-size-case-studies}c). This U-shaped behavior across molecular weight demonstrates that LLMs do not fail uniformly, but rather exhibit a narrow window of partial plausibility shaped more by dataset and pretraining biases than by true spectral reasoning.

Finally, our CoT analysis reveals that the internal reasoning traces are verbose (200–435 words) yet unreliable: DBE computations almost never match ground truth, formula counts are nearly always incorrect, and contradiction rates exceed 70\% across all models. This mirrors the structural results: LLMs can fluently emulate expert analytical language, including references to neutral losses, ring patterns, or DBE formulas, but these textual elements are not mechanistically linked to the final molecular predictions. The superficially detailed explanations thus mask a lack of chemical grounding and often contain inconsistencies within the reasoning trace itself or between the CoT and the generated SMILES. CoT-guided reasoning, in this setting, enhances interpretability only at the surface level while offering little improvement in correctness.

Taken together, these findings reveal that text-only LLMs lack (1) grounding in fragmentation physics, (2) mechanisms to enforce formula or valency constraints, and (3) the ability to integrate local reasoning steps into globally coherent molecular graphs. These limitations persist across model families, sizes, and molecular-weight regimes, and become particularly stark when contrasted with spectra-trained models like DiffMS. Progress in MS-to-structure prediction will therefore require integrating LLMs with domain-grounded components such as spectral encoders, chemically constrained decoding, hybrid symbolic–differentiable reasoning modules, and multimodal pretraining that couples textual reasoning with molecular graphs and spectra. While CoT prompting enables LLMs to produce fluent, expert-like narratives, it does not confer the mechanistic understanding required for reliable mass spectral structure elucidation. Substantial architectural and multimodal advances will be necessary to bridge this gap.

\section{Limitations and Conclusion}
\label{sec:conclusion}

This work establishes a systematic benchmark for evaluating LLMs on MS-to-SMILES prediction under CoT prompting. While all models reliably follow the structured \texttt{<think>} and \texttt{<answer>} format and often generate syntactically valid SMILES, our results reveal clear limitations in chemical fidelity and spectral grounding.

Taken together, our findings reveal that the limitations of zero-shot LLMs arise from a combination of weak chemical priors, unreliable reasoning, and structural ambiguity inherent to MS/MS data. Because none of the models receive task-specific training, their behavior reflects only the generic chemical knowledge acquired during pretraining, which manifests in uniformly poor chemical metrics: DBE accuracy remains below 4.2\%, formula correctness is effectively zero, and exact-match accuracy is zero across all models and molecular-weight bins. Although CoT prompting increases reasoning transparency, it does not improve reasoning accuracy, DBE computations are correct in only 3.9--4.2\% of cases, formula statements are almost always wrong, and contradiction rates exceed 72\%, indicating that the seeming “expert-like" explanations are largely disconnected from the generated structures. Even when examining molecular-weight effects, where all models perform best on mid-sized molecules (200-400~Da), the improvements are marginal: Tanimoto similarity still hovers around 0.10 and MCES overlap remains partial, with performance degrading sharply for larger molecules due to combinatorial explosion and for very small molecules due to sparse fragment information. Altogether, these patterns show that zero-shot LLMs, even when scaffolded with CoT, lack the mechanistic grounding required to perform reliable spectral-structure inversion.

Despite these constraints, our benchmark provides a clear and standardized view
of how contemporary LLMs behave on mass spectral reasoning tasks. The findings
suggest that achieving chemically grounded MS-to-structure prediction will
require integrating LLMs with chemical constraints, spectral encoders, or
physics-informed priors. Bridging these components with natural-language
reasoning represents a promising path toward reliable, interpretable molecular
analysis.

\section{Abbreviations}

\begin{description}
    \item[MS] Mass spectrometry
    \item[MS/MS] Tandem mass spectrometry
    \item[FP] Fingerprints
    \item[DBE] Double bond equivalent
    \item[LLM] Large language model
    \item[CoT] Chain-of-thought
    \item[GRU] Gated recurrent unit
    \item[SVM] Supported vector machine
    \item[CNN] Convolution neural network
    \item[RNN] Recurrent neural network
    \item[LSTM] Long short-term memory model
    \item[GNN] Graph neural network
    \item[GDM] Graph diffusion model
    \item[RL] Reinforcement learning
    \item[MCES] Minimum graph dissimilarity
    \item[SMILES] Simplified Molecular Input Line Entry System
    \item[MW] Molecular weight
    \item[SOTA] State of the art
\end{description}

\section{Declarations}

\subsection{Availability of data and materials}

All benchmark experiments in this study are conducted on the publicly available MassSpecGym dataset \cite{bushuiev2024massspecgym}.  
Our CoT prompting templates and evaluation code for all models are openly available in GitHub repository: \url{https://github.com/airscker/ms-cot-benchmark}

\subsection{Competing interests}

The authors declare that they have no competing interests.

\subsection{Funding}

This research did not receive any specific grant from funding agencies in the public, commercial, or not-for-profit sectors.

\subsection{Authors' contributions}

Yufeng Wang and Wei Lu designed and implemented the benchmarking framework, carried out the experiments, and performed the data analysis.  
Lin Liu provided domain expertise in mass spectrometry and chemistry, contributed to research discussion, assisted in interpreting the results and manuscript revision.  
Hao Xu and Haibin Ling conceived and supervised the overall project, guided the methodological development, and contributed to the interpretation and presentation of findings.  
All authors discussed the results, contributed to manuscript writing and revision, and approved the final version of the paper.

\subsection{Acknowledgements}

We thank the developers of MassSpecGym and prior MS-to-structure methods (including MSNovelist, MassGenie, MIST, DiffMS, and MADGEN) for making their datasets and implementations publicly available, which made this benchmarking study possible.  
We are also grateful to colleagues and lab members for helpful discussions and feedback during the development of this work.

\subsection{Authors' information}
Yufeng Wang and Lu Wei contributed equally to this work.

\subsubsection*{Authors and Affiliations}
\textbf{Department of Computer Science, Stony Brook University, 100 Nicolls Road, Stony Brook, NY 11794, United States.}\\Yufeng Wang, Wei Lu, Haibin Ling

\noindent\textbf{Department of Chemistry, Stanford University, 450 Jane Stanford Way, Stanford, CA 94305, United States.}\\Lin Liu

\noindent\textbf{Department of Medicine, Brigham and Women's Hospital, Harvard Medical School, 77 Avenue Louis Pasteur, Boston MA 02115, United States.}\\Hao Xu

\noindent\textbf{Department of Artificial Intelligence, Westlake University, No.600 Dunyu Road, Hangzhou 310030, Zhejiang Province, China.}\\Haibin Ling

\bibliography{reference}

\begin{thebibliography}{10}

\bibitem{wolfender2018accelerating}
Jean-Luc Wolfender, Jean-Marc Nuzillard, Justin~JJ Van Der~Hooft, Jean-Hugues
  Renault, and Samuel Bertrand.
\newblock Accelerating metabolite identification in natural product research:
  toward an ideal combination of liquid chromatography--high-resolution tandem
  mass spectrometry and nmr profiling, in silico databases, and chemometrics.
\newblock {\em Analytical Chemistry}, 91(1):704--742, 2018.

\bibitem{patti2012metabolomics}
Gary~J Patti, Oscar Yanes, and Gary Siuzdak.
\newblock Metabolomics: the apogee of the omics trilogy.
\newblock {\em Nature reviews Molecular cell biology}, 13(4):263--269, 2012.

\bibitem{theodoridis2011mass}
Georgios Theodoridis, Helen~G Gika, and Ian~D Wilson.
\newblock Mass spectrometry-based holistic analytical approaches for metabolite
  profiling in systems biology studies.
\newblock {\em Mass spectrometry reviews}, 30(5):884--906, 2011.

\bibitem{stein2012mass}
Stephen Stein.
\newblock Mass spectral reference libraries: an ever-expanding resource for
  chemical identification.
\newblock {\em Analytical Chemistry}, 2012.

\bibitem{duhrkop2019sirius}
Kai D{\"u}hrkop, Markus Fleischauer, Marcus Ludwig, Alexander~A Aksenov,
  Alexey~V Melnik, Marvin Meusel, Pieter~C Dorrestein, Juho Rousu, and
  Sebastian B{\"o}cker.
\newblock Sirius 4: a rapid tool for turning tandem mass spectra into
  metabolite structure information.
\newblock {\em Nature methods}, 16(4):299--302, 2019.

\bibitem{stravs2022msnovelist}
Michael~A Stravs, Kai D{\"u}hrkop, Sebastian B{\"o}cker, and Nicola Zamboni.
\newblock Msnovelist: de novo structure generation from mass spectra.
\newblock {\em Nature Methods}, 19(7):865--870, 2022.

\bibitem{shrivastava2021massgenie}
Aditya~Divyakant Shrivastava, Neil Swainston, Soumitra Samanta, Ivayla Roberts,
  Marina Wright~Muelas, and Douglas~B Kell.
\newblock Massgenie: a transformer-based deep learning method for identifying
  small molecules from their mass spectra.
\newblock {\em Biomolecules}, 11(12):1793, 2021.

\bibitem{litsa2023end}
Eleni~E Litsa, Vijil Chenthamarakshan, Payel Das, and Lydia~E Kavraki.
\newblock An end-to-end deep learning framework for translating mass spectra to
  de-novo molecules.
\newblock {\em Communications Chemistry}, 6(1):132, 2023.

\bibitem{goldman2023annotating}
Samuel Goldman, Jeremy Wohlwend, Martin Stra{\v{z}}ar, Guy Haroush, Ramnik~J
  Xavier, and Connor~W Coley.
\newblock Annotating metabolite mass spectra with domain-inspired chemical
  formula transformers.
\newblock {\em Nature Machine Intelligence}, 5(9):965--979, 2023.

\bibitem{bohde2025diffms}
Montgomery Bohde, Mrunali Manjrekar, Runzhong Wang, Shuiwang Ji, and Connor~W
  Coley.
\newblock Diffms: Diffusion generation of molecules conditioned on mass
  spectra.
\newblock {\em arXiv preprint arXiv:2502.09571}, 2025.

\bibitem{wang2025madgen}
Yinkai Wang, Xiaohui Chen, Liping Liu, and Soha Hassoun.
\newblock Madgen--mass-spec attends to de novo molecular generation.
\newblock {\em arXiv preprint arXiv:2501.01950}, 2025.

\bibitem{yang2024structural}
Yiming Yang, Shuang Sun, Shuyuan Yang, Qin Yang, Xinqiong Lu, Xiaohao Wang,
  Quan Yu, Xinming Huo, and Xiang Qian.
\newblock Structural annotation of unknown molecules in a miniaturized mass
  spectrometer based on a transformer enabled fragment tree method.
\newblock {\em Communications Chemistry}, 7(1):109, 2024.

\bibitem{honda2019smiles}
Shion Honda, Shoi Shi, and Hiroki~R Ueda.
\newblock Smiles transformer: Pre-trained molecular fingerprint for low data
  drug discovery.
\newblock {\em arXiv preprint arXiv:1911.04738}, 2019.

\bibitem{priyadarsini2024self}
Indra Priyadarsini, Seiji Takeda, Lisa Hamada, Emilio~Vital Brazil, Eduardo
  Soares, and Hajime Shinohara.
\newblock Self-bart: A transformer-based molecular representation model using
  selfies.
\newblock {\em arXiv preprint arXiv:2410.12348}, 2024.

\bibitem{litsa2021spec2mol}
Eleni Litsa, Vijil Chenthamarakshan, Payel Das, and Lydia Kavraki.
\newblock Spec2mol: An end-to-end deep learning framework for translating ms/ms
  spectra to de-novo molecules.
\newblock {\em ChemRxiv, Organic Chemistry}, 2021.

\bibitem{chen2021molecular}
Fangying Chen, Junyoung Park, and Jinkyoo Park.
\newblock A molecular hyper-message passing network with functional group
  information.
\newblock {\em arXiv preprint arXiv:2106.01028}, 2021.

\bibitem{chen2024molecular}
Junwu Chen and Philippe Schwaller.
\newblock Molecular hypergraph neural networks.
\newblock {\em The Journal of Chemical Physics}, 160(14), 2024.

\bibitem{le2020neuraldecipher}
Tuan Le, Robin Winter, Frank No{\'e}, and Djork-Arn{\'e} Clevert.
\newblock Neuraldecipher--reverse-engineering extended-connectivity
  fingerprints (ecfps) to their molecular structures.
\newblock {\em Chemical science}, 11(38):10378--10389, 2020.

\bibitem{wei2022chain}
Jason Wei, Xuezhi Wang, Dale Schuurmans, Maarten Bosma, Fei Xia, Ed~Chi, Quoc~V
  Le, Denny Zhou, et~al.
\newblock Chain-of-thought prompting elicits reasoning in large language
  models.
\newblock {\em Advances in neural information processing systems},
  35:24824--24837, 2022.

\bibitem{ouyang2022training}
Long Ouyang, Jeffrey Wu, Xu~Jiang, Diogo Almeida, Carroll Wainwright, Pamela
  Mishkin, Chong Zhang, Sandhini Agarwal, Katarina Slama, Alex Ray, et~al.
\newblock Training language models to follow instructions with human feedback.
\newblock {\em Advances in neural information processing systems},
  35:27730--27744, 2022.

\bibitem{qiuphenomenal}
Linlu Qiu, Liwei Jiang, Ximing Lu, Melanie Sclar, Valentina Pyatkin, Chandra
  Bhagavatula, Bailin Wang, Yoon Kim, Yejin Choi, Nouha Dziri, et~al.
\newblock Phenomenal yet puzzling: Testing inductive reasoning capabilities of
  language models with hypothesis refinement.
\newblock In {\em The Twelfth International Conference on Learning
  Representations}.

\bibitem{lewkowycz2022solving}
Aitor Lewkowycz, Anders Andreassen, David Dohan, Ethan Dyer, Henryk
  Michalewski, Vinay Ramasesh, Ambrose Slone, Cem Anil, Imanol Schlag, Theo
  Gutman-Solo, et~al.
\newblock Solving quantitative reasoning problems with language models.
\newblock {\em Advances in Neural Information Processing Systems},
  35:3843--3857, 2022.

\bibitem{shao2024deepseekmath}
Zhihong Shao, Peiyi Wang, Qihao Zhu, Runxin Xu, Junxiao Song, Xiao Bi, Haowei
  Zhang, Mingchuan Zhang, YK~Li, Y~Wu, et~al.
\newblock Deepseekmath: Pushing the limits of mathematical reasoning in open
  language models.
\newblock {\em arXiv preprint arXiv:2402.03300}, 2024.

\bibitem{li2023chain}
Xingxuan Li, Ruochen Zhao, Yew~Ken Chia, Bosheng Ding, Shafiq Joty, Soujanya
  Poria, and Lidong Bing.
\newblock Chain-of-knowledge: Grounding large language models via dynamic
  knowledge adapting over heterogeneous sources.
\newblock {\em arXiv preprint arXiv:2305.13269}, 2023.

\bibitem{yao2023react}
Shunyu Yao, Jeffrey Zhao, Dian Yu, Nan Du, Izhak Shafran, Karthik Narasimhan,
  and Yuan Cao.
\newblock React: Synergizing reasoning and acting in language models.
\newblock In {\em International Conference on Learning Representations (ICLR)},
  2023.

\bibitem{xu2025towards}
Fengli Xu, Qianyue Hao, Zefang Zong, Jingwei Wang, Yunke Zhang, Jingyi Wang,
  Xiaochong Lan, Jiahui Gong, Tianjian Ouyang, Fanjin Meng, et~al.
\newblock Towards large reasoning models: A survey of reinforced reasoning with
  large language models.
\newblock {\em arXiv preprint arXiv:2501.09686}, 2025.

\bibitem{yao2023tree}
Shunyu Yao, Dian Yu, Jeffrey Zhao, Izhak Shafran, Tom Griffiths, Yuan Cao, and
  Karthik Narasimhan.
\newblock Tree of thoughts: Deliberate problem solving with large language
  models.
\newblock {\em Advances in neural information processing systems},
  36:11809--11822, 2023.

\bibitem{pei2023biot5}
Qizhi Pei, Wei Zhang, Jinhua Zhu, Kehan Wu, Kaiyuan Gao, Lijun Wu, Yingce Xia,
  and Rui Yan.
\newblock Biot5: Enriching cross-modal integration in biology with chemical
  knowledge and natural language associations.
\newblock {\em arXiv preprint arXiv:2310.07276}, 2023.

\bibitem{edwards2021text2mol}
Carl Edwards, ChengXiang Zhai, and Heng Ji.
\newblock Text2mol: Cross-modal molecule retrieval with natural language
  queries.
\newblock In {\em Proceedings of the 2021 Conference on Empirical Methods in
  Natural Language Processing}, pages 595--607, 2021.

\bibitem{ghugare2023searching}
Raj Ghugare, Santiago Miret, Adriana Hugessen, Mariano Phielipp, and Glen
  Berseth.
\newblock Searching for high-value molecules using reinforcement learning and
  transformers.
\newblock {\em arXiv preprint arXiv:2310.02902}, 2023.

\bibitem{bushuiev2024massspecgym}
Roman Bushuiev, Anton Bushuiev, Niek de~Jonge, Adamo Young, Fleming Kretschmer,
  Raman Samusevich, Janne Heirman, Fei Wang, Luke Zhang, Kai D{\"u}hrkop,
  et~al.
\newblock Massspecgym: A benchmark for the discovery and identification of
  molecules.
\newblock {\em Advances in Neural Information Processing Systems},
  37:110010--110027, 2024.

\bibitem{newell2019international}
David~B Newell, Eite Tiesinga, et~al.
\newblock The international system of units (si).
\newblock {\em NIST Special Publication}, 330(1), 2019.

\end{thebibliography}
\bibliographystyle{unsrt}

\section{Appendix}
\subsection{Domain-Specific Prompt Template}
\begin{center}
  \begin{minipage}{\linewidth}
    \begin{lstlisting}
You are an expert in analyzing Mass Spectra (MS) to deduce the corresponding SMILES. Given the MS m/z values: <mzs>, intensities: <intensities>, and molecular formula: <formula>, instrument: <instrument>, adduct (Ionization Method): <adduct>, and collision energy: <collision_energy> eV, your task is to analyze the spectrum and propose plausible SMILES representation candidates of the molecule. Please provide your analysis in <think> block and the most probable SMILES in <answer> block with the following template:
<think>
1. Formula and DBE Analysis:
* Formula: <formula>
* Double Bond Equivalents (DBE): <DBE> (Calculated as C - H/2 + N/2 + 1)
* Initial structural implications from DBE: (e.g., presence of rings, double/triple bonds)
2. Key Peak Identification and Initial Fragmentation:
* Base Peak: m/z <base_peak_mz> with intensity <base_peak_intensity>. Potential stable fragment/substructure: <substructure_formula_from_base_peak>.
3. Neutral Loss Analysis from Significant Peaks:
* From M+ (or <source_peak_mz>): Loss of <mass_A> (m/z <molecular_ion_mz> -> m/z <fragment_A_mz>) suggests loss of <neutral_fragment_A_formula_or_structure>.
* From m/z <source_peak_B_mz>: Loss of <mass_B> (m/z <source_peak_B_mz> -> m/z <fragment_B_mz>) suggests loss of <neutral_fragment_B_formula_or_structure>.
* Common neutral losses observed (e.g., H2O, CO, NH3, C2H4): <list_common_losses_and_implications>.
4. Fragment Ion Relationship and Substructure Assembly:
* Peak at m/z <fragment_X_mz> likely arises from <parent_ion_mz_for_X> via <loss_or_rearrangement_for_X>. Proposed substructure for <fragment_X_mz>: <subformula_X>.
* Peak at m/z <fragment_Y_mz> and its relation to <fragment_X_mz> or other peaks.
* Hypothesize connections between identified substructures (<subformula_A>, <subformula_X>, etc.) consistent with <formula> and DBE.
5. Structure Elucidation and SMILES Hypothesis:
* Synthesize fragment information, neutral losses, and DBE to propose a consistent chemical structure.
* Consider plausible fragmentation mechanisms for the observed spectrum.
</think>
<answer>
Final 10 SMILES Proposals: <smiles_1>,<smiles_2>,...,<smiles_10>
</answer>
    \end{lstlisting}
  \end{minipage}
\end{center}

\subsection{Demo Predictions}

\begin{table}[!h]
\centering
\begin{tabular}{@{}ccl@{}}
\toprule
\multicolumn{3}{c}{\textbf{Small molecule example ($<200$~Da)}} \\
\midrule
\textbf{Index} & \textbf{Tanimoto Similarity} & \textbf{SMILES} \\
\midrule
1  & 0.0000 & CCCCN \\
2  & 0.0000 & CC(C)CN \\
3  & 0.0000 & CC(C)(C)N \\
4  & 0.0000 & CNCCC \\
5  & 0.0000 & C(C)C(C)N \\
6  & 0.0667 & CNC(C)C \\
7  & 0.0000 & NCC(C)C \\
8  & 0.0000 & NC(C)(C)C \\
9  & 0.0000 & CCNCC \\
10 & 0.0000 & C(CN)CC \\
\midrule
\multicolumn{3}{c}{\textbf{Medium molecule example ($200-400$~Da)}} \\
\midrule
\textbf{Index} & \textbf{Tanimoto Similarity} & \textbf{SMILES} \\
\midrule
1  & 0.4545 & NCC(C1=CC=C(O)C=C1)C(=O)O \\
2  & 0.3611 & OC1=CC=CC(C(C(=O)O)N)=C1 \\
3  & 0.3333 & NCC(C2=CC=CC(O)=C2)C(=O)O \\
4  & 0.2703 & CC1=CC=C(O)C=C1C(=O)N \\
5  & 0.5152 & OC1=CC=CC=C1CC(N)C(=O)O \\
6  & 1.0000 & NC(Cc1ccc(O)cc1)C(=O)O \\
7  & 0.3333 & NCC(C3=CC(O)=CC=C3)C(=O)O \\
8  & 0.3158 & NCC(C4=CC=CC=C4O)C(=O)O \\
9  & 1.0000 & O=C(O)C(N)Cc1ccc(O)cc1 \\
10 & 0.5152 & O=C(O)C(N)Cc1ccccc1O \\
\midrule
\multicolumn{3}{c}{\textbf{Large molecule example ($>400$~Da)}} \\
\midrule
\textbf{Index} & \textbf{Tanimoto Similarity} & \textbf{SMILES} \\
\midrule
1  & 0.1154 & CC(C)(C)c1ccc(C(=O)c2ccc(C(C)(C)C)cc2)cc1 \\
2  & 0.0893 & CCCCCCCCCc1ccc(O)cc1 \\
3  & 0.0847 & CC(C)CCCCCCc1ccc(O)cc1 \\
4  & 0.1154 & CC(C)C1=CC=C(C(=O)C2=CC=C(C(C)C)C=C2)C=C1 \\
5  & 0.1053 & CC(C)CC(C)C1=CC(=O)C=CC1=O \\
6  & 0.1207 & CCCCCCCCC1CC(O)CCC1 \\
7  & 0.0806 & CCCCC(C)Cc1ccccc1O \\
8  & 0.1311 & CCCCCCCCCCCC1OC(=O)C=CC1 \\
9  & 0.1017 & CCCCCC1=CC=CC(=O)C1=O \\
10 & 0.0806 & CC(C)CCCC1CCC(C(C)C)CO1 \\
\bottomrule
\end{tabular}
\caption{\textbf{10 candidate structures for three representative spectra.}
For each small, medium, and large molecule example, we report the rank,
fingerprint-based Tanimoto similarity to the ground-truth structure, and the
candidate SMILES string. The medium-sized case contains exact matches
(Tanimoto = 1.00) that are not ranked first, while the small and large cases
show uniformly low similarity, illustrating different failure modes of the
model across molecular-weight regimes.}
\label{tab:case-study-candidates}
\end{table}

\end{document}